\documentclass[10pt, journal,fleqn]{IEEEtran}  % Comment this line out if you need a4paper
\pdfoutput=1
\IEEEoverridecommandlockouts     % This command is only needed if  % you want to use the \thanks command
\usepackage{color,amsmath,amstext,amsfonts,amssymb, mathrsfs,mathtools}
\usepackage{graphicx,algorithm,algorithmic}
\usepackage{tikz}
\usepackage{setspace}
\usepackage{array} 
\usepackage{booktabs}  
\usepackage{bbm}
\usepackage{wasysym}
\usepackage{textcomp}
\usepackage{hyperref} 
\usepackage{multirow,graphicx}
\usepackage[sort,compress]{cite}
\usepackage[justification=centering]{caption}
\usepackage[normalem]{ulem}

\renewcommand{\thesubsubsection}{{\it \Alph{subsection}.\arabic{subsubsection}.}}

\newcounter{l1}
\newcounter{l2}
\newcounter{l3}
\newcommand{\bdotlist}{\begin{list}{$\bullet$}{}}
\newcommand{\bboxlist}{\begin{list}{$\Box$}{}}
\newcommand{\bbboxlist}{\begin{list}{\raisebox{.005in}{{\tiny $\blacksquare$ \ \ }}}{}}
\newcommand{\bdashlist}{\begin{list}{$-$}{} }
\newcommand{\blist}{\begin{list}{}{} }
\newcommand{\barablist}{\begin{list}{\arabic{l1}}{\usecounter{l1}}}
\newcommand{\balphlist}{\begin{list}{(\alph{l2})}{\usecounter{l2}}}
\newcommand{\bAlphlist}{\begin{list}{\Alph{l2}.}{\usecounter{l2}}}
\newcommand{\bdiamlist}{\begin{list}{$\diamond$}{}}
\newcommand{\bromalist}{\begin{list}{(\roman{l3})}{\usecounter{l3}}}

\newcommand{\beq}{\begin{equation}}
\newcommand{\eeq}{\end{equation}}

\definecolor{darkgreen}{RGB}{0,0,0}
\newcommand{\green}[1]{{\color{darkgreen}#1}}

\title{\LARGE \bf
Graph Learning for Cognitive Digital Twins in Manufacturing Systems
}

\author{Trier Mortlock$^\ddag$, Deepan Muthirayan$^\dagger$, Shih-Yuan Yu$^\dagger$, Pramod P. Khargonekar$^\dagger$, Mohammad A. Al Faruque$^{\dagger\ddag}$\\
Department of Electrical Engineering and Computer Science$^\dagger$\\ Department of Mechanical and Aerospace Engineering$^\ddag$\\ University of California, Irvine, California, USA\\ (tmortloc, dmuthira, shihyuay, pramod.khargonekar, alfaruqu)@uci.edu
}

\begin{document}
\maketitle
%\markboth{}{} 

\thispagestyle{empty}
\pagestyle{empty}

\begin{abstract}
Future manufacturing requires complex systems that connect simulation platforms and virtualization with physical data from industrial processes. Digital twins incorporate a physical twin, a digital twin, and the connection between the two. Benefits of using digital twins, especially in manufacturing, are abundant as they can increase efficiency across an entire manufacturing life-cycle. The digital twin concept has become increasingly sophisticated and capable over time, enabled by rises in many technologies. In this paper, we detail the cognitive digital twin as the next stage of advancement of a digital twin that will help realize the vision of Industry 4.0. Cognitive digital twins will allow enterprises to creatively, effectively, and efficiently exploit implicit knowledge drawn from the experience of existing manufacturing systems. They also enable more autonomous decisions and control, while improving the performance across the enterprise (at scale). This paper presents graph learning as one potential pathway towards enabling cognitive functionalities in manufacturing digital twins. %Graph learning leverages the inherent graph structures that can be observed throughout manufacturing systems and the ability to learn across non-Euclidean data. 
A novel approach to realize cognitive digital twins in the product design stage of manufacturing that utilizes graph learning is presented.  
\end{abstract}
\let\thefootnote\relax\footnote{\scriptsize{This work has been submitted to the IEEE for possible publication. Copyright may be transferred without notice, after which this version may no longer be accessible.}}


\section{Future of Manufacturing}
\label{sec:fut-of-manufacturing}

%\textcolor{orange}{Opening: added stress on \textit{computing} as per requested for the journal version} 
The evolution of manufacturing can be presently defined by four major transformations: (i) the industrial revolution during the 18th and 19th centuries; (ii) mass production in the first half of the 20th century; (iii) information technology-based automation of production in the second half of the 20th century; and (iv) the ongoing fourth industrial revolution. This fourth industrial revolution has many monikers, namely, Industry 4.0, Smart Manufacturing, connected industries --- as part of Society 5.0 in Japan, Made in China 2025, etc. This wave of future visions of manufacturing is often hinged on harnessing the power of computing in manufacturing systems. Most of these visions aspire to bring together wireless (and wired) communications, smart sensors, cyber-physical systems, internet-of-things (IoT) \cite{jazdi2014cyber}, advanced robotics~\cite{Bahrin2016INDUSTRY4A}, additive manufacturing, simulation and high-performance computing, advanced data analytics, machine learning and artificial intelligence,  cloud computing, and cybersecurity \cite{industry4-alfaruque}. The goal of the fourth industrial revolution is to achieve personalized, affordable, efficient, resilient, adaptive, and sustainable products and production across distributed factories and supply chains. 
%\cite{jazdi2014cyber, atzori2010internet, holler2015machine}
%nolfi2000evolutionary

\begin{figure*}[!ht]
    \centering
    \includegraphics[width=0.93\linewidth]{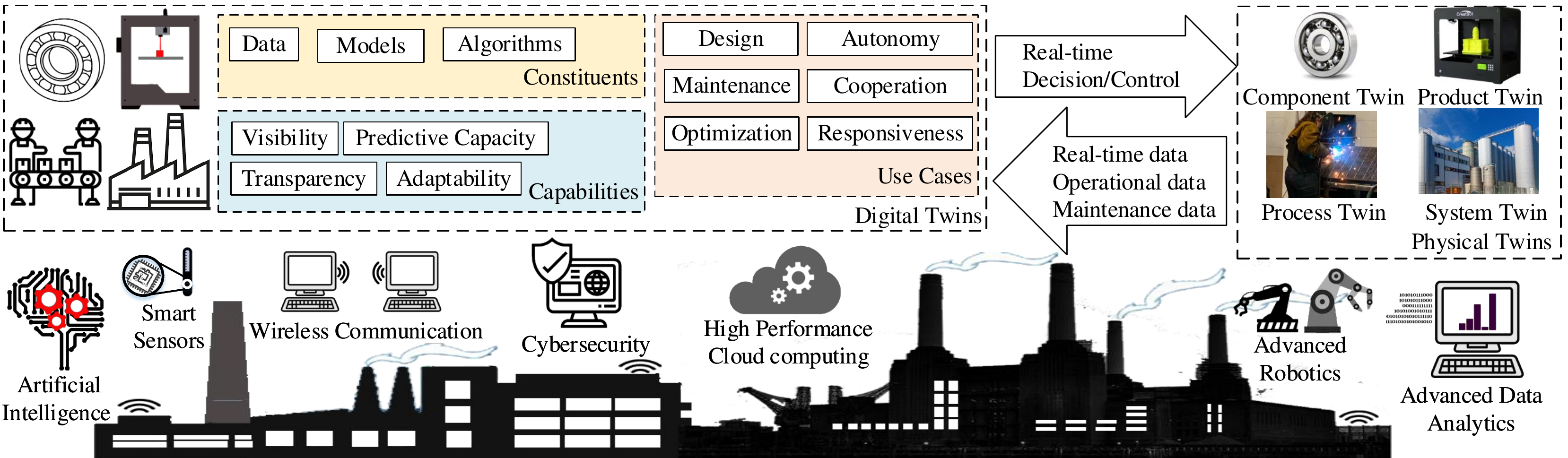}
    \caption{Digital Twin in the context of cyber-physical production systems \cite{alf2021cognitive}.}
    \label{fig:DT-I4}
\end{figure*}

%\textcolor{orange}{Digital Twins Intro Paragraph} 
Digital twins are one of the many technologies that will help shape the future of manufacturing. Digital twins integrate the cyber and physical worlds --- allowing for seamless communication between digital models and real, in-operation systems. Digital twins have been shown to provide distinct advantages in manufacturing \cite{alfaruqudigitaltwin}. % \cite{alfaruqudigitaltwin,chhetri2019quilt} % and can impact a wide range of applications such as design, planning, optimization, and maintenance of products, processes and systems. %The  potential of digital twins has given rise to many exciting and challenging research questions. Although 
Digital twins leverage many auxiliary technologies and systems such as modeling and simulation, IoT sensors, standards and interoperability among digital technologies, computing, and data from different stages of the product lifecycle. These enabling technologies that constitute increasingly sophisticated and powerful digital twins are illustrated in Fig.~\ref{fig:DT-I4}.

%\textcolor{orange}{Relevance of Digital Twins Paragraph} 
The deployment of digital twins in manufacturing can have an impact throughout the entire product life cycle --- including, but not limited to, product design and optimization, testing, production system design and operation, supply chain management and control, prognostics, maintenance, aftermarket services, and cybersecurity. As with many modern research fields in computing, the applications of digital twins in manufacturing continues to rapidly progress and increase. Gartner has listed the digital twin as one of the top ten technology trends for 2019 and the years to come \cite{gartner2019}. Furthermore, according to recent reports, the digital twins' market size is projected to grow from US\$3.1 billion in 2020 to US\$48.2 billion in value by 2026 \cite{marketsandmarkets2020}. Historically, the aerospace and automotive industries have been dominant users of digital twin technologies; however, other manufacturing sectors are also beginning to leverage the many advantages that digital twins can offer \cite{marketsandmarkets2020}. %\cite{marketsandmarkets2020,deloitte2020}
Concurrently, digital twins have given rise to many exciting and challenging research questions.

%\textcolor{orange}{Paragraph on learning for cognition} 
%Machine learning methods can be one of many ways to enable functionalities in digital twins. It is without question that machine learning technologies are becoming pervasive throughout modern and emerging technologies [ADD CITATIONS]. Leveraging large amounts of data, researchers and companies alike have begun to tackle more complex problems where learning methods can improve results. Furthermore, enhancements in the computing power of hardware enables an even wider usage of learning implementations [ADD Citation]. In this paper, one specific type of machine learning, graph learning, is chosen to demonstrate the proposed digital twins for manufacturing systems.

In this paper, we extend upon the cognitive digital twin framework introduced in \cite{alf2021cognitive}. Specifically, we propose graph learning frameworks and algorithms as a potential pathway for cognitive digital twins. Graph learning is a subset of machine learning that focuses on algorithms and methods involving data that can be structured as graphs. Graph learning techniques, which have gained increasing interest in recent years \cite{wu2020comprehensive}, have many benefits in modeling non-Euclidean data that has inherent graph structures. The pervasive nature of graphs in manufacturing systems can be observed at all scales and stages --- from process and fabrication flow graphs to high-level organization supply chains. Hence, graph learning can be leveraged in future Industry 4.0 systems. Furthermore, these methods can be used to realize the vision of cognition in digital twins. We make the important clarification that graph learning is not the only solution to enabling cognition in digital twins, nor does it completely enable all levels of cognition across a system. It is simply just yet another computational method that takes advantage of data-driven modelling that can be used in systems moving towards further autonomy.

The main contribution of this work is a novel, graph learning framework for enabling cognition in digital twins. The framework we propose is a query-based framework that can potentially answer any type of query. The framework involves three main steps namely, (i) the {\it Graph Formation} step, which can aggregate data from diverse data types and from diverse products in a graph representation that can be used for reasoning or solving a specific problem, (ii) {\it Graph Operations} that process the graph further to select the key features, and (iii) the {\it Learning Objective} that specifies the problem to be solved to answer the query. This framework can leverage both physical and virtual data simultaneously, improve the core operations of manufacturing systems, and address some of the open research gaps in the field. %We envision that this work provides another step towards Industry 4.0 by providing a pathway for more cognitive manufacturing system designs.

The remainder of this paper is structured as follows: Section \ref{sec:digital-twin} introduces digital twins and their importance in Industry 4.0; in Section \ref{sec:cognitive-digital-twin}, the concept of cognitive digital twins is detailed; Section \ref{sec:graphlearning} describes concepts of graph learning and its application in manufacturing along with a general graph learning framework for digital twins with illustration for the product design stage; and Section \ref{sec:conclusion} gives concluding remarks.

\section{Digital Twins and Manufacturing}
\label{sec:digital-twin}

%\textcolor{orange}{The only real change here may just be a paragraph at the end specifying graph strucutres that are present in digital twins. }

The term \textit{Digital Twin} was first used in 2002 by John Vickers of the National Aeronautics and Space Administration (NASA). During a later 2010 report, Vickers also gave the first formal definition of the digital twin, specified for air vehicles, as \textit{an integrated multi-physics, multi-scale, probabilistic simulation of an as-built vehicle or system that uses the best available physical models, sensor updates, fleet history, etc., to mirror the life of its corresponding flying twin} \cite{negri2017review}.
The most basic and concise definition of a digital twin can be stated as in \cite{hultgren2020concept}: ``\textit{A digital twin has a digital or a virtual part, a physical part and a connection between them.}''

Since its inception, the digital twin concept has evolved and expanded to various products, processes, and domains. Digital twins have been proposed for many applications or use cases such as predicting real-time behavior, monitoring, decision support, planning, production optimization, and control \cite{kritzinger2018digital, tao2018digital}. 
%\cite{kritzinger2018digital, tao2018digital, shao2019digital}
\green{ The concept of digital twins in cyber-physical production systems is studied in \cite{uhlemann2017digital}, however strictly for data acquisition. In \cite{ding2019defining}, a digital twin-based cyber-physical production system is examined, placing specific emphasis on the interconnection between the physical and cyber-shop floors. Our focus encompasses these works in a larger view of cyber-physical manufacturing (see Fig. \ref{fig:DT-I4}) while providing a flexible framework for enabling cognitive functions throughout the system’s life cycle.}

Fig. \ref{fig:DT-I4} illustrates the digital twin in the context of a cyber-physical production systems (CPPS) and the overarching use cases (or functions). The top part of the figure illustrates the digital twin ecosystem: (i) major building blocks and constituents of digital twins (data, models, and algorithms); (ii) capabilities digital twins can enable (visibility, transparency, predictive capacity, and adaptability); and (iii) the various use cases ranging from design to autonomy and cooperation. The bottom part of the figure shows the key enabling technologies of this ecosystem. The top-right side of the figure depicts the physical manufacturing system. CPPS related data (e.g., operational data and maintenance data) are collected in real-time and provided to the digital twins. Digital twins in response provide real-time feedback (e.g., decision and control) to the physical manufacturing system. This provides real-time two-way seamless communication between the physical manufacturing system and the corresponding digital twin.   

The use cases of digital twins span the entire life cycle of a product: product design and optimization, testing, production system design and operation, supply chain management and control, prognostics, maintenance, aftermarket services, cyber-security (Fig. \ref{fig:DT-I4}). For example, the digital twin of a component or a product can be used to simplify and streamline the design process by enabling virtual testing of the specifications to ensure that the product meets the standards (verify) and the performance requirements (validate) \cite{tao2018digital}. The digital twin of a product can be used to detect the early onset of faults (predictive maintenance), help diagnose the fault, and provide customized solutions for performance optimization, maintenance, and compliance \cite{tao2018digital}. A manufacturing company can offer new services based on digital twins for optimizing the performance of the product during operation. Thus, computing through digital twins provides organizations with opportunities for offering new products and services. The digital twin of a production system can be used to optimize the system for throughput (performance), reduce waste (efficiency), improve quality \cite{tao2018digital}. Digital twins can also enable processes involved in CPPS to be adaptable and responsive to disruptive events \cite{lim2019state}. 

\begin{figure}[ht]
    \centering
    \includegraphics[width=1.0\linewidth]{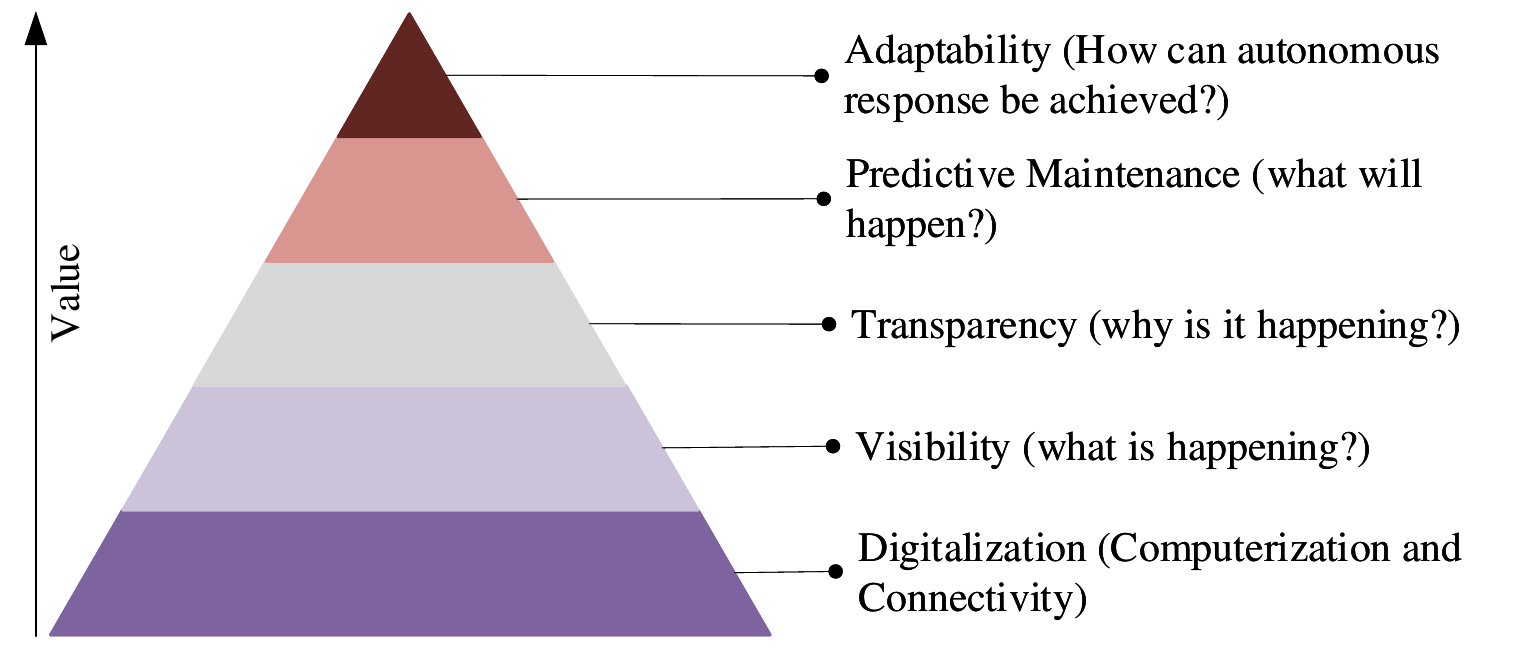}
    \caption{Pathway to Industry 4.0 \cite{alf2021cognitive}.} 
    \label{fig:pathway-I4}
\end{figure}

% \begin{figure*}[!ht]
%     \centering
%     \includegraphics[width=0.8\linewidth]{CDT-ManSystems/figures/diff.pdf}
%     \caption{Standard digital twin and cognitive digital twin \cite{alf2021cognitive}.}
%     \label{fig:digitaltwin}
% \end{figure*}

In our previous paper we envisioned that cognitive digital twins are the key for the various stages of the Industry 4.0 development path (proposed in \cite{ignatov2019global}): visibility (what is happening), transparency (why is it happening), predictive capacity (what will happen), and adaptability (how can an autonomous response be achieved) (see Fig. \ref{fig:pathway-I4}, which is an adaptation of Fig. 3 in \cite{ignatov2019global}). Visibility is vital for assessing the shop floor changes and operating conditions for the overall system's adaptability and responsiveness. This will impact decisions in the production pipeline, allowing the system to adapt quickly and effectively, therefore reducing downtime and costs. Transparency and predictive capacity are essential for understanding and inferring how to respond, and thus directly affect manufacturing performance under changing circumstances. Finally, adaptability enables the system to respond to a changing situation by itself with self-corrections based on feedback, instead of a human-in-the-loop making the decisions. When this happens across interacting physical processes, the system can achieve seamless cooperation in how it responds and operates.  %Digital twins can --- and will --- play a huge role in each of these successive stages in progress in Industry 4.0.

%\textcolor{orange}{%Many aspects of digital twins have inherent graph structures. %Graph learning is detailed explicitly in Section \ref{sec:graphlearning}, however we take a moment to stress the wide applicability of graphs throughout digital twins. As stated above, the digital twin of a system can have many goals, but it is primarily tasked with providing an accurate representation of a physical system. 
%Many complex systems have increasingly connected components that enable operation in a variety of dynamic environments. Graphs can simply represent [info here]. Graphs can also be abstracted to represent [more info here] (functions of a system), or unforeseen relations or impossible mathematical connections). In the case of manufacturing, these systems have a wide range and scope, but nonetheless have similar properties. More sentences here on relation to networks and graphs. [Need to find some more papers to cite for this section. Could we create a figure based on this paragraph to illustrate our ideas?]}

There are many different algorithms and implementations that may be used to realize these discussed capabilities in digital twins (\green{e.g., Neural Networks, Dynamic Bayesian Networks, Hidden Markov Models, Finite Element Analysis, and Deep Learning  \cite{lim2019state}}). Graph learning, another example, can enhance visibility by learning relationships between systems and subsystems --- creating a higher-level abstracted vision of an entire system. Graph representations can also provide more explainable decisions than traditional machine learning algorithms as relations between components of a system can be mapped and visually interpreted. Also, graph learning can help in prediction tasks --- for example, if a tool in a manufacturing plant breaks down or requires repair, the graph structure can enable informed predictions on other tools that may need preventative repairs or reinforcements. Large scale graphs can enable higher levels of autonomy in digital twins by allowing abstraction through a holistic view of a system, much like how humans perceive their environment. {\it We posit that graph learning, further detailed in Section \ref{sec:graphlearning}, provides a pathway for cognition in digital twins}.
% Graphs based on large amounts of data can enable a higher level of aunontomy in digital twinsalso help a digital twin move towards a higher level of autonomy by allowing abstraction at a holistic view of a system, much like humans do in their thought processes.  
\section{Cognitive Digital Twin} % section 
\label{sec:cognitive-digital-twin}
%\textcolor{orange}{I can't any think of a good way to add graph learning in this section}
The definition of \textit{cognitive digital twin} is inspired by major advances in cognitive science, machine learning, and artificial intelligence. 
Neisser’s classic definition of cognition~\cite{solso2005cognitive} includes “\textit{all the processes by which the sensory input is transformed, reduced, elaborated, stored, recovered and used ...}”. 
Fundamental aspects of cognition include attention (selective focus), perception (forming useful precepts from raw sensory data), memory (encoding and retrieval of knowledge), reasoning (drawing inferences from observations, beliefs, and models), learning (from experiences, observations, and teachers), problem-solving (achieving goals), knowledge representation, etc.

\begin{figure}
    \centering
    \includegraphics[width=1.\linewidth]{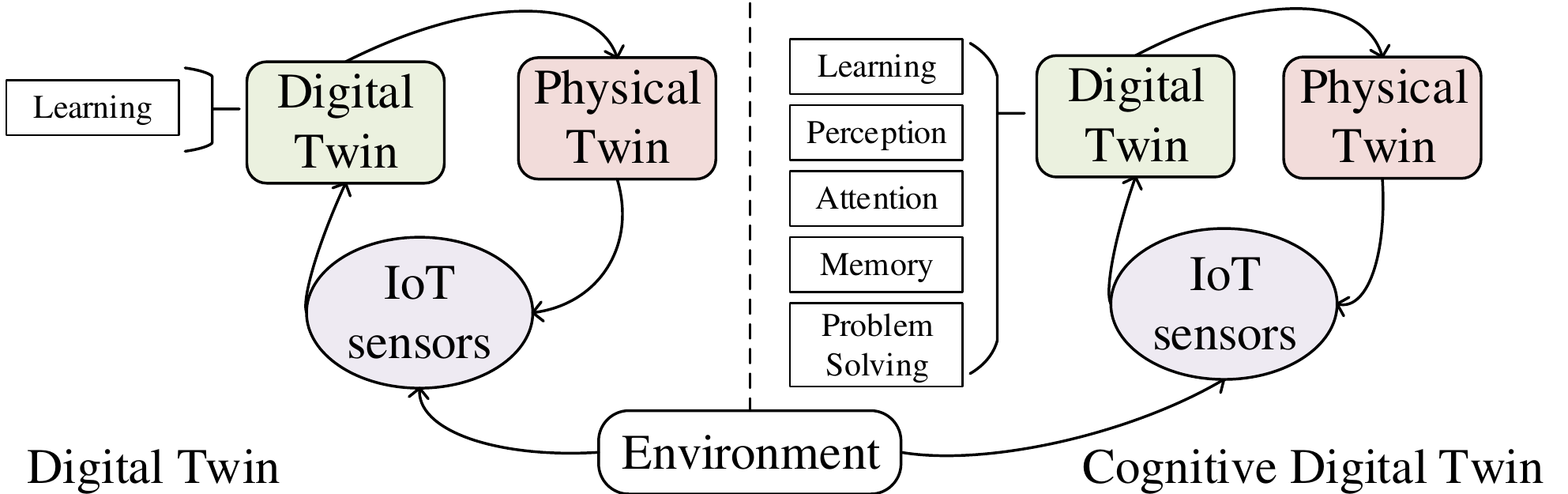}
    \caption{Standard digital twin and cognitive digital twin \cite{alf2021cognitive}.}
    \label{fig:digitaltwin}
\end{figure}

The standard view of the digital twin and the conceptual framework of the cognitive digital twin proposed in \cite{alf2021cognitive} are depicted in Fig. \ref{fig:digitaltwin}. The digital twin on the left of Fig. \ref{fig:digitaltwin} is the standard digital twin, which has a digital part, a corresponding physical part, and a connection between them. This version of the digital twin has the ability to learn. The digital twin we propose is shown on the right of Fig. \ref{fig:digitaltwin}, which in addition to having the ability to learn, is endowed with the other elements of cognition such as perception, attention, memory, reasoning, problem-solving, etc. In the following, we describe these capabilities in the context of a digital twin. 

%maybe adjust or remove this figure
% \begin{figure*}[!ht]
%     \centering
%     \includegraphics[width=0.9\linewidth]{CDT-ManSystems/figures/cognitive_dt.pdf}\vspace{-0.5em}
%     \caption{Aspects of human cognition and its exemplary realization to the proposed enabling components in Digital Twins}
%     \label{fig:cognitivedt}\vspace{-1.5em}
% \end{figure*}

\subsection{Cognitive Capabilities}
\label{sec:cognitive-digital-twin-capabilities}

\subsubsection{Perception} in cognitive psychology could be defined as {\it the organization, identification, and interpretation of sensation to form a mental representation} \cite{schacter2011psychology}. We extend this definition to define perception in cognitive digital twin as {\it the process of forming useful representations of data related to the physical twin and its physical environment for further processing}. It is well established that machine learning techniques are less effective in learning representations of high dimensional and large data volume \cite{lecun2015deep}. Since CPPS (and IoT) generate multi-modal, high-dimensional, large volumes of data, we posit that perception is a key cognitive capability to form useful {\it precepts} upon which further cognitive processing can occur in a digital twin. Perception in a digital twin will enable visibility in manufacturing systems.

\subsubsection{Attention} can be viewed as the allocation of limited resources or a selection mechanism \cite{posner2011cognitive}. %\cite{oberauer2019working, posner2011cognitive, cohen2014cognitive}
We adopt the latter view and define attention in a cognitive digital twin as {\it the process of focusing selectively on a task or a goal or certain sensory information either by intent or driven by environmental signals and circumstances}. Attention can be perceptual or non-perceptual and controlled or otherwise (see \cite{oberauer2019working} for a detailed taxonomy of attention). Attention enables focus on the essential information from the raw sensor data and memory. So, it can simplify and improve the process of perception and decision making in a cognitive digital twin. Attention will help monitor or select a task to focus on, paving the way for autonomy in manufacturing systems. 

\subsubsection{Memory} we define memory in a cognitive digital twin adopting the view of memory provided in \cite{mcdermott2018memory}: {\it is a single process that reflects a number of different abilities: holding information briefly while working with it (working memory), remembering episodes of the physical twin’s life (episodic memory), and knowledge of facts of the environment and its interaction with the physical twin (semantic memory), where remembering includes the steps: encoding information (learning it, by perceiving it and relating it to past knowledge), storing it (maintaining it over time), and then retrieving it (accessing the information when needed)}. Thus, memory (both working memory and the remembered episodes and knowledge), are an essential ingredient for the algorithms complementing the digital twin to autonomously control the physical processes related to the various stages of a physical twin because memory allows the algorithm to remember the context and additionally allows the digital twin to leverage past knowledge.

\subsubsection{Reasoning} in cognitive psychology can be broadly defined as the ``process of drawing meaningful conclusions for informing problem-solving or decision making'' \cite{sternberg2004nature}. Reasoning can be broadly classified under deduction, induction, and probabilistic reasoning \cite{st1993cognitive}. {\it Thinking} and {\it reasoning} are cornerstones of human intelligence and so have been extensively studied in cognitive psychology \cite{baron2000thinking,sternberg2004nature}. %\cite{baron2000thinking,sternberg2004nature,johnson2006we,stenning2012human}
We define reasoning in cognitive digital twins adopting the definition proposed in \cite{johnson2010mental}: {\it drawing conclusions consistent with a starting point — a perception of the physical twin and its environment, a set of assertions, a memory, or some mixture of them}. Thus, reasoning directly impacts understanding (transparency) and is central to decision making (autonomy). 

\subsubsection{Problem-solving} we define problem-solving in cognitive digital twin as {\it the process of finding a solution for a given problem or achieving a given goal from a starting point}. Thus, problem-solving is central to decision making and autonomy.

\subsubsection{Learning} we define learning in cognitive digital twin as {\it the process of transforming experience of the physical twin into reusable knowledge for a new experience}. 
Hence, learning is essential for adaptability (or autonomy) and responsiveness of the physical system that the digital twin represents and becomes a key ingredient for intelligence in digital twins. 
\section{Graph Learning for  Cognitive Digital Twins}
\label{sec:graphlearning}
Data driven models based on artificial intelligence and machine learning have become increasingly popular for enabling some cognitive capabilities in digital twins at various levels of manufacturing systems \cite{abburu2020cognitwin}. In this section, we present graph learning as just one of many solutions that can be used to enable cognition in digital twins. %\cite{bannat2010artificial,jiang2020human,abburu2020cognitwin,eirinakis2020enhancing}
%Machine learning methods using graphs offer flexibility and adaptability in multiple different cognitive capabilities which will be shown in Section \ref{sec:mandesign}.
%\textcolor{orange}{Insert some advantages of graph learning basically explaining why we chose it as our method to implement cognition in digital twins --- stay high-level though because we will cover graph advantages specifically in the next section.}
In the subsequent subsections, we briefly provide preliminaries on graph learning along with motivation for the use of graphs in manufacturing systems. In Section \ref{sec:mandesign2}, we present specialized graph learning frameworks and discuss how they can enable cognitive capabilities in digital twins and offer potential solutions for important research challenges within the field of manufacturing systems.

\begin{figure}[h!]
\centering
  \includegraphics[scale = 0.35]{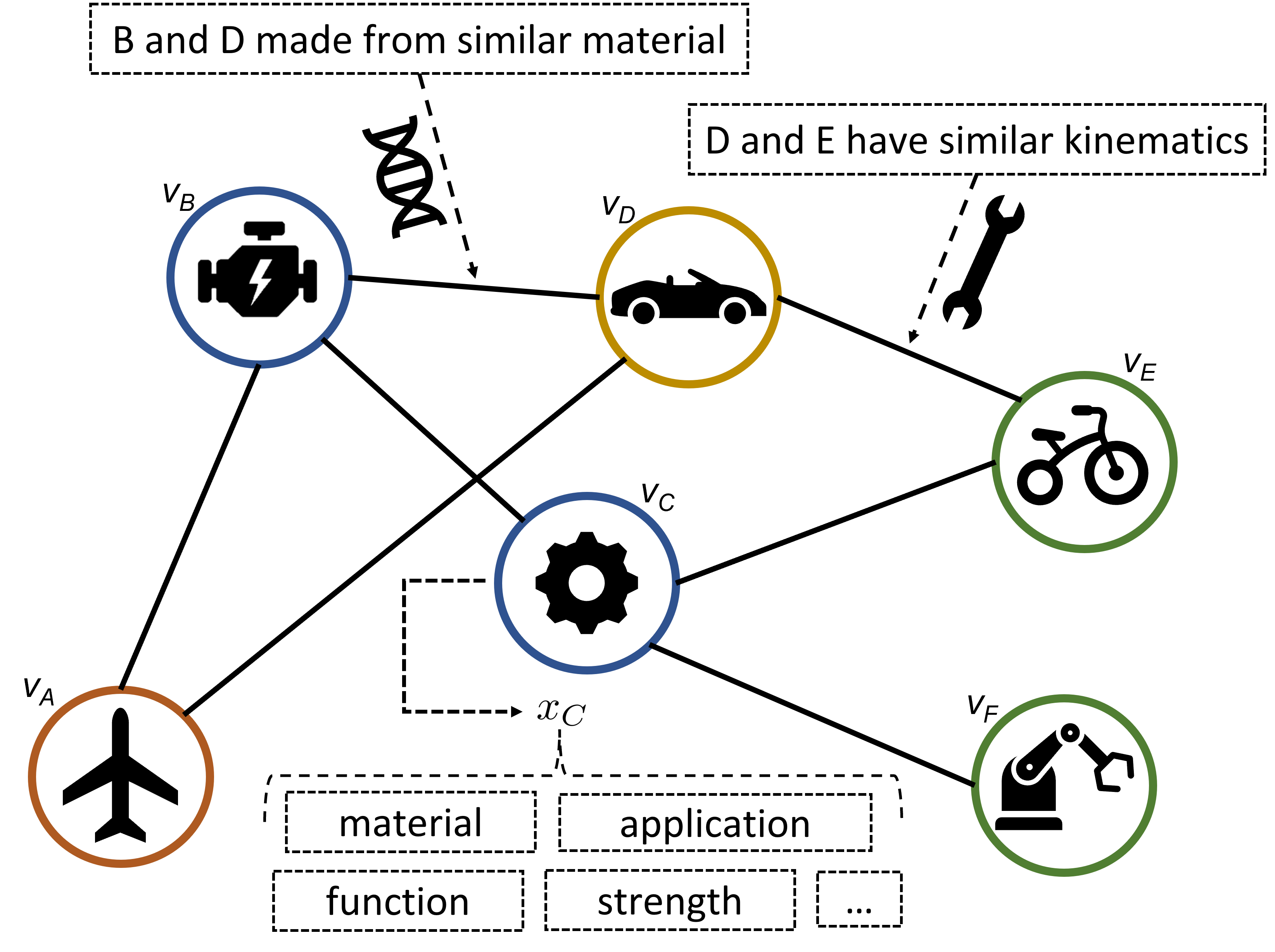}
  \caption{A generic undirected graph formed in the context of product design.}
  \label{fig:Graph_Pre}
\end{figure}

\begin{figure}[h!]
\centering
  \includegraphics[scale = 0.30]{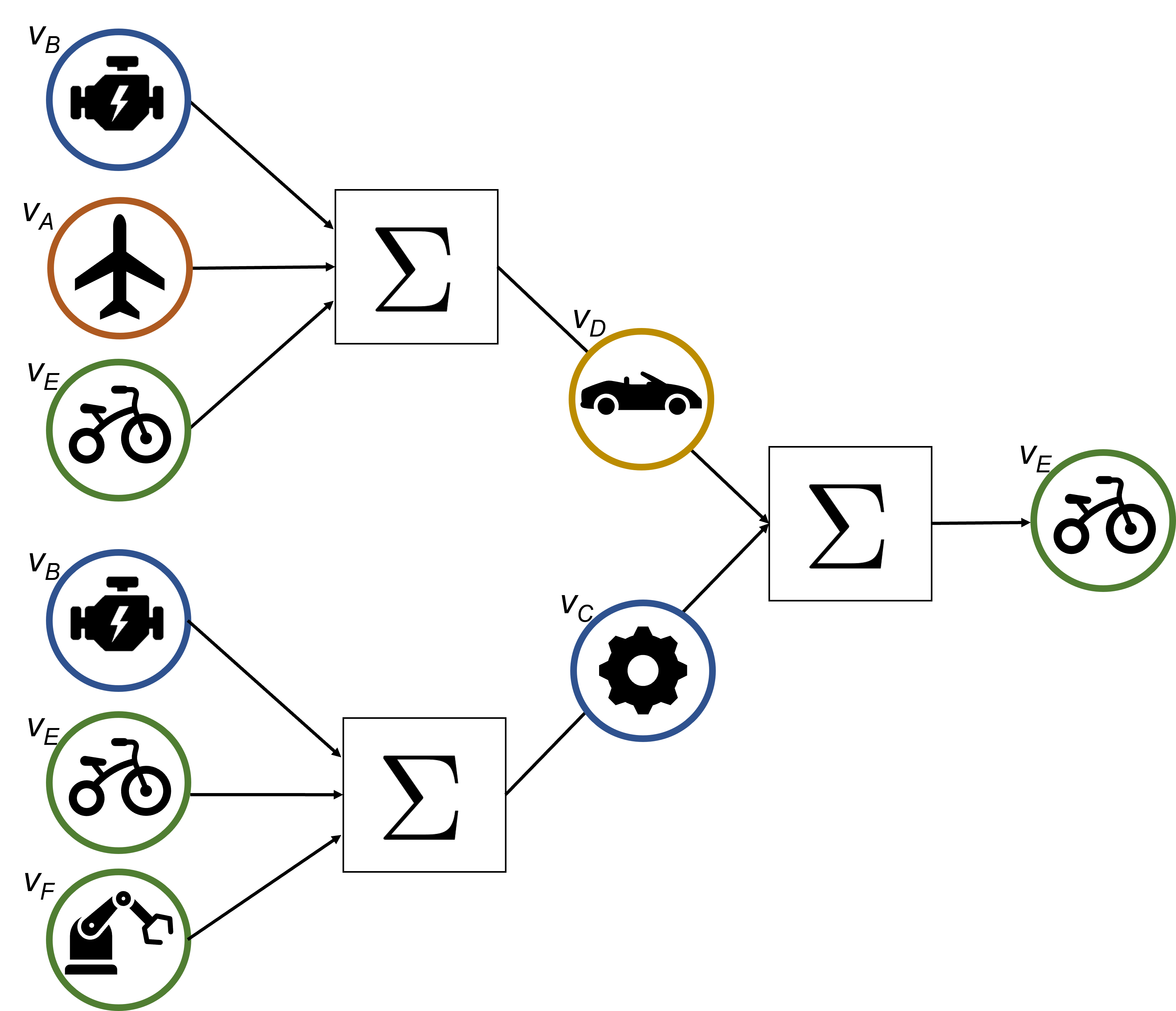}
  \caption{Neighborhood node aggregation of node $E$.}
  \label{fig:Neighbors}
\end{figure}

\begin{table*}[t]
\centering
\caption{Related Works of Graphs Learning in Manufacturing}
\label{table:graph}
\begin{tabular}{|c|c|c|}
\hline
\multicolumn{1}{|c|}{\textbf{Manufacturing Stage}} & 
\multicolumn{1}{|c|}{\textbf{Graph Function}} & 
\multicolumn{1}{|c|}{\textbf{References}} \\ \hline

\multirow{3}{*}{Design} & perform automation engineering tasks using functional lifting & \cite{wan2018future}\\ \cline{2-3}
 & design of distributed circuits & \cite{zhang2019circuit}\\ \cline{2-3}
 & unsupervised 3D shape retrieval & \cite{nie2020multi} \\ \hline
 
\multirow{2}{*}{Optimization} & optimize scheduling performance of flexible manufacturing systems & \cite{hu2020petri} \\ \cline{2-3}
& find the optimum scheduling policy for job-shop problems & \cite{park2021learning} \\ \hline

\multirow{4}{*}{Monitoring} & detect and isolate faulty components in industrial systems & \cite{khorasgani2019fault} \\ \cline{2-3}
%\hline %\\ %\hline %\cline{2-4}
 & improve product failure prediction & \cite{kang2020product} \\ \cline{2-3}
 & share multi-level manufacturing knowledge within a system & \cite{he2019manufacturing} \\ \cline{2-3}
 & predict the remaining useful life estimation of industrial equipment  & \cite{narwariya2020graph} \\ \hline %\cline{2-4}
 %& text & 1\\ \cline{2-4}
% \hline
%\label{table:graph} 
\end{tabular}
\end{table*}
\subsection{Preliminaries on Graph Learning}
\label{ssec:prelims}
%Paragraph with some background on graphs learning - include seminal works and important citations
% The reader is referred to \cite{wu2020comprehensive} as a survey paper for further recent trends in graph learning. %\cite{hamilton2017representation,wu2020comprehensive}

%paragraph with Graph basics and accompanying figure(s)
We define a graph with $n$ nodes as, $\mathcal{G} = \{\bar{V},\bar{A}\}$, where $\bar{V}= [v_1, v_2, ... v_n]$ is the matrix of the node embedding in the graph, $v_i$ is the node embedding for node $i$. The edges of a graph are represented by the adjacency matrix, $\bar{A}$, where each value $a_{ij}$ corresponds to the relation between nodes $i$ and $j$. The feature matrix of the graph, $\bar{X}$ is composed of each nodes feature vectors, $x_i$, which can hold unique data pertaining to each node. Fig. \ref{fig:Graph_Pre} shows a basic graph of six nodes with various connections. This example is set in a manufacturing design setting where each node is a product, and the edges are possible connections between products. Additionally, some examples of features of product $C$ are also shown. 

Graphs can be classified as (i) {\it directed} or {\it undirected} and (ii) {\it weighted} or {\it unweighted}. Directed graphs are graphs where the edges represent relations that are uni directional. The opposite is the undirected graph where the edge is a bi-directional relation. Weighted graphs are graphs where the edges have an associated weight that specifies the relative importance of the relation. The concepts we present later hold for graphs of any type. %However, for our analysis, the concepts presented hold broadly without making distinctions between the exact type of graph chosen.

Graph learning broadly includes two steps: the formation of the graph and finding a lower dimensional representation of the graph. Graph neural networks are a widely popular and successful deep learning technique for generating lower dimensional representation of graph structure data. The first \textit{Graph Neural Network} (GNN) model was developed in \cite{scarselli2008graph}, and since then, numerous graph approaches have been proposed such as \textit{Recurrent Graph Neural Networks} (RecGNNs), \textit{Convolutional Graph Neural Networks} (ConvGNNs), \textit{Graph Autoencoders} (GAEs) and \textit{Spatio-temporal Graph Neural Networks} (STGNNs)~\cite{wu2020comprehensive}. 
One of the key techniques contributing to graph learning's rise in popularity is notably the Graph Convolutional Network (GCN), developed by \cite{kipf2016semi} in 2016.

Graph neural networks typically include the following two operations: (i) message passing and (ii) node embedding. Message passing is the function that aggregates features from a node's neighbors. This operation updates the features of nodes with the information from their respective neighbors and its current feature values. Multiple message passing operations will then result in a final node feature that is a function of information from nodes across the graph.
% \begin{equation}
%     a_v^{(k)} = \textsc{AGGREGATE}^{(k)}(\{h_u^{(k-1)}: u \in N(v)\}) \nonumber 
% \end{equation}
% \begin{equation}
%     h_v^{(k)} = \textsc{COMBINE}^{(k)}(h_v^{(k-1)}, a_v^{(k)} )\nonumber 
% \end{equation}
% \begin{equation}
%     h^{(k)}_g = \textsc{GRAPH\_READOUT}(\{h_v^{(k)}: v \in V\})\nonumber 
% \end{equation}
Figure \ref{fig:Neighbors} shows one such instance of message passing for the graph in Fig. \ref{fig:Graph_Pre}. The node embedding operation encodes the final feature vector as a lower dimensional representation. These two operations form the key parts of most general graph neural network approach. %From a high level this process is depicted in Fig. \ref{fig:Embed}, with a simple graph which highlights the embedding of nodes $a$ and $b$. The goal in this illustration is for the nodes in the embedded space to maintain information and relations while transforming into the lower dimensional $z$-space, or mathematically:
% The goal in this operation is for the nodes in the embedded space to maintain information and relations while transforming into the lower dimensional $z$-space, or mathematically:
% \begin{center}
%     $z^a = Enc(x^a)\quad,$
% \end{center}
% \begin{center}
%      $Similarity (a,b) = (z^a)^T(z^b) = (x^a)^T(x^b) \quad.$
% \end{center}
%Although graphs can be much more complex, these basic modelling techniques still hold. Also fundamentals of permutation equivariance and permutation in-variance remain intact throughout the graph neural network to conserve the integrity of the mathematical operations performed. 
% \begin{figure}[h!]
% \centering
%   \includegraphics[scale = 0.6]{figures2/Embed.png}
%   \caption{Embedding depiction of two nodes into the $z$-space.}
%   \label{fig:Embed}
% \end{figure}

%\textcolor{orange}{Do we need a paragraph on some challenges of graph learning here? Or maybe this commentary should go right before we detail message passing and node embedding so it shows why the techniques are valuable.}
% Figures we might wanted to consider adding:
% - generic graph structure to help illustrate basics
% - maybe one graph in action --- doing some learning 

\subsection{Graph Learning in Manufacturing}
%Paragraph with advantages of using graphs.
Graph learning has some fundamental advantages, specifically: (i) modeling of non-Euclidean data; (ii) expressing data in an insightful manner to understand relationships between entities of a system; (iii) abstracting data for higher-level reasoning tasks; (iv) capturing dependencies between the different model instances of a system. Since graphs are prevalent in manufacturing systems (see Section \ref{sec:digital-twin}), it follows that graph learning can provide significant advantages in manufacturing. In Table \ref{table:graph}, works that illustrate the use of graph learning in various phases of manufacturing systems are listed. In \cite{wan2018future,zhang2019circuit,nie2020multi}, it is shown that graphs of similar or related products can be used to leverage information sharing and save time and money for verification, validation, and testing in the design stage. In \cite{hu2020petri,park2021learning}, it is shown that graph learning can be used to optimize scheduling performance during operation. In monitoring systems, graph learning has been shown to improve detection and prediction of failures in both products and equipment \cite{khorasgani2019fault,kang2020product,narwariya2020graph}. While these demonstrations highlight the opportunities that lie for graph learning in manufacturing systems, the graph learning systems that these works develop are restricted to specific applications. In the following sections, we present a novel and general graph learning framework for any digital twin application. We focus on the design stage of manufacturing to illustrate our ideas.

\begin{figure*}[!ht]
    \centering
    \includegraphics[width=0.9\linewidth]{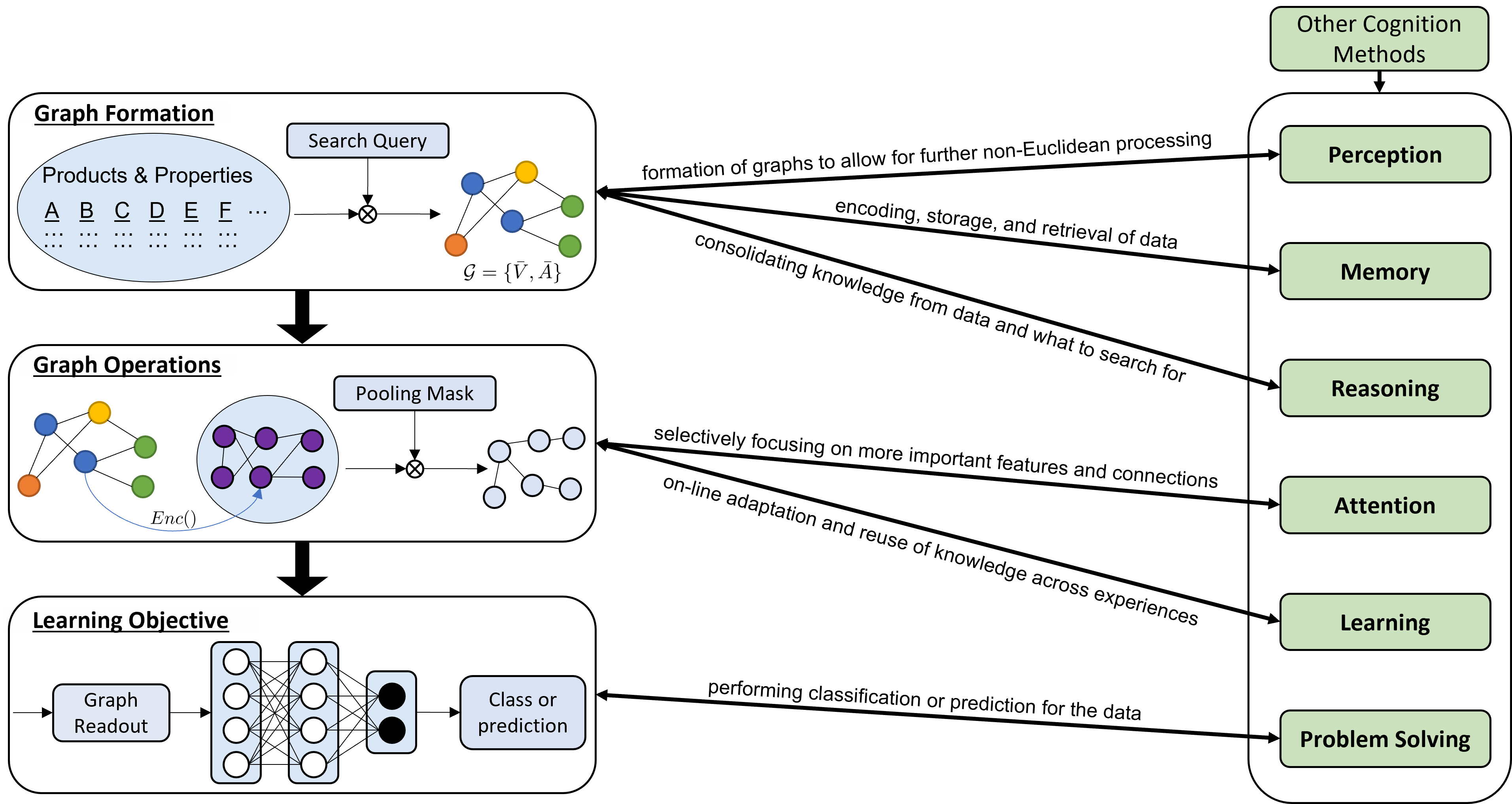}\vspace{-0.5em}
    \caption{An example of a cognitive digital twin using graph learning during the product design stage of manufacturing. \\
    The graph learning framework is composed of three steps: \textbf{(i) Graph Formation}: products and their properties are filtered by a user defined design principle query to form a graph of the products and their relations; \textbf{(ii) Graph Operations}: higher level abstractions or relations are obtained from the generated graph; and \textbf{(iii) Learning Objective}: defines the problem to be solved (e.g., classification, prediction) for answering the query and the general metrics and specifications to optimize and fine-tune the model.}
    \label{fig:graph-cogdt}\vspace{-1.5em}
    \vspace{1.5em}
\end{figure*}

\subsection{Product Design}
\label{sec:mandesign}
\green{Benefits of cognition can be realized throughout manufacturing, like shown in \cite{zaeh2010holistic}. They propose a cognitive production planning and control system that it is able to react and autonomously adapt its production planning process to increase manufacturing efficiency. They employ cognitive functions of perception, knowledge, and planning within their architecture. Cognitive IoT is another application that can be used in manufacturing systems as sensors themselves can exercise cognitive capabilities \cite{ploennigs2017materializing}.} While cognition can impact various stages of Industry 4.0 and related technologies, here we focus on the product design stage to illustrate our key ideas. %\cite{jiani2017, ahmed2017, zaeh2010holistic}
The potential impact of this focus is far-reaching --- advancements in product design can lead to more robust designs, cost efficient research and development, and overall more streamlined designs. The product design stage of manufacturing can be defined as steps starting from the idea generation of a product to the physical testing, verification, and validation of the product before manufacturing. Current practices in this domain often target their efforts in two separate streams, one that focuses on physical data and the other that focuses on virtual data. Thus both do not take advantage of the benefits a digital twin can offer \cite{tao2018digital2}. %\cite{tao2018digital2,tao2019digital} 
%Narrowing our focus once more, we highlight the cognitive opportunities in \textit{searching} throughout product design.
In the following subsections, we discuss the application of graph learning to the {\it search} operation of the design phase. We discuss the benefits that arise and comment on the limitations and future research directions.
% 4: https://link.springer.com/content/pdf/10.1007/s00170-017-0233-1.pdf
% 5: https://www.tandfonline.com/doi/pdf/10.1080/00207543.2018.1443229

\subsection{The Search Operation}
The \textit{search} operations in the product design stage can be defined as processes that involve transferring useful knowledge and information from other products or systems that can aid in the design of the product under development. 
%By leveraging data currently available, more robust and cost-effective product development can be performed if these relationships can be determined and mapped by a product’s digital twin. This is a nontrivial task that involves sharing knowledge across domains, embedding complex data throughout digital twins, and building cognitive capabilities using the data
Currently, researchers and engineers spend a large amount of time building a new product, process, or system without leveraging the knowledge that already exists for a similar or somewhat closer product, process, or system. %\cite{reusing-design, arango1993process}
For example, there are resources such as GrabCad \cite{grabcad}, which contains computer-aided design (CAD) models of many engineering systems, that can be used as a base design for creating new models. These CAD models are part of the digital twin of the product that goes through the manufacturing process to be converted into the physical twin. Various methods have been proposed for exploring 3D models (a partial digital twin model defining the physical twin geometry and manufacturing information). For example, authors in \cite{3Dmodelsearch} created a 3D search engine that utilizes the spherical harmonics descriptor for acquiring the signature vector and use the Euclidean distance among these vectors to find a similar polygon model. Authors in \cite{VESPER} utilized the shape similarity metric on 3D models to enable the discovery of parts and assemblies existing in the company's database. Authors in \cite{3DPDF} utilized a similarity measure based on various attributes (such as name, description, etc.) to calculate a score for discovering CAD models in repositories. While these research efforts focused on searching and discovering similar 3D models based on geometry information, the digital twin models comprise more than just geometry information of the physical twins and thus can be used for searching and responding to more diverse search queries. {\it We posit that graph learning techniques by virtue of their capability to learn general relationships are critical to building a general purpose product search engine.}

\subsection{Graph Learning Framework}
\label{sec:mandesign2}
%Applications of graph learning in product designs have been explored in a variety of fields, e.g. circuit design and 3D shape retrieval \cite{zhang2019circuit,nie2020multi}. \sout{In \cite{wan2018future}, the authors present a specialized graph learning framework that uses the concept of functional lifting for future automation engineering in the product design space. They define automation engineering as the design, creation, development and management of production systems in factories, process plants, and supply chains that realize the production of products. Automated functional lifting refers to the strategy of inferring functions of products based on abstractions of common engineering data. By leveraging graph learning, the authors were able to show accurate classification of product types using engineering data artifacts. Despite the progress of these works in applying graph learning to the product design stage, they do not capitalize on the ability to share and learn knowledge while searching across different domains.}

In Fig. \ref{fig:graph-cogdt}, we portray the graph learning framework for enabling cognition in digital twins. Cognitive functions allow humans to transfer their knowledge and experiences from one domain to a completely different domain. While this is a big challenge for current machine learning methodologies, graph learning is a potential machine learning technique that can bridge this gap. %{\it We propose graph learning as a potential methodology to enable human-like knowledge transfer in digital twins}. 
In Fig. \ref{fig:graph-cogdt}, the functional steps of the graph learning framework are shown on the left and the cognitive functions it enables are shown on the right. 

The framework we propose is a query-based framework. The query can be either any user generated query or an automated query from the digital twin. The query can potentially cover a range of tasks or problems like finding functionally similar products or generating configuration for a new set of product specifications, or even generating novel insights for new products. Thus, the scope of this framework is very broad. The graph learning framework will allow the digital twin to answer such higher level queries by identifying complex product to product or product to sub-component relationships and solving problems by processing this information. In the \textbf{Graph Formation} step, data of products and their properties are mined and organized as a graph based on the query, allowing them to be processed further. The constructed graph captures the essential relations and provides a powerful abstraction or representation that can enable reasoning or problem solving using graph operations to answer the query. The \textbf{Graph Operations} step, can model very general and complex mathematical functions. Graph operations can further aggregate information and form condensed representations as shown in the figure, which is useful for generating the final decision or insight to answer the query. Graph neural networks, in principle, can model any mathematical function, and so are a potential approach for this step. Lastly, the \textbf{Learning Objective} specifies the problem to be solved for answering the query and the general metrics and specifications to optimize and fine-tune the model for solving this problem. Specifying the correct metrics for training and verifiability are critical to ensure the digital twin can effectively and reliably optimize the model for the given problem (e.g. classification, prediction). Accurate problem specification is also critical for generating the best response to the query. This step's output and subsequent feedback can also be used to iteratively refine and improve the model. 

By leveraging this framework, digital twins can learn to organize information from diverse domains by their complex relations based on the query (which can be a search query). %through the detection of functional similarities across non-Euclidean data.
While physical testing is not completely substitutable, this method can make the design process more efficient by allowing the designer to leverage prior knowledge. For example, a designer targeting a specific design objective can initiate their query, and the framework can enable aggregation of knowledge from various domains to generate new insights for the design of the product. This will also enable digital twins to leverage physical and virtual data simultaneously, and find improved solutions for the specific queries, addressing a key gap that we alluded to in Section \ref{sec:mandesign}. We also envision that future cognitive digital twins will be able to make these queries autonomously. We note that full cognition in manufacturing will leverage a combination of other methods, as indicated in the top right of Fig. \ref{fig:graph-cogdt}, and that graph learning is one of the key methodologies. \green{In the next section, we showcase a real-world application that illustrates some functionalities of the proposed framework.}   %This novel approach for cognitive digital twins addresses priors limitations in sharing knowledge across domains and incorporating higher-lever cognitive tasks for design. 

\green{ 
\section{Use Case}
\label{sec:app}
In this section, we review experiments and results of an application to illustrate the proposed methodology. In a previous work, our group presented a specialized graph learning framework that uses the concept of functional lifting for future automation engineering in the product design space \cite{wan2018future}. Automation engineering in this context can be defined as automation of the \textit{design, creation, development and management of production systems in factories, process plants, and supply chains that realize the production of products}. Automated functional lifting refers to the strategy of inferring functions of products based on abstractions of common engineering data. %By leveraging graph learning, the authors were able to show accurate classification of product types using engineering data artifacts. 
%\red{More specific on classification here. simplify this paragraph. merge with next paragraph maybe.}
In these experiments, the goal is to produce accurate classification labels for different product types using engineering data artifacts. This classification enables more efficient search operation within and across different products with similar functionalities. To realize these operations, a graph learning framework with a structural graph convolutional neural network (SGCNN) is proposed.

%the specific graph learning model that is deployed to perform the actions of the cognitive digital twin in the presented setting. %move to next section
% Give more context on this classification, it enables searching for products of similar properties. To realize this the convolutional is proposed to realize the cognitive function of [specifics]. \red{This example illustrates the search operation within manufacturing product design, this example concretely illustrates benefits that can be realized via graph learning in the domain of future manufacturing designs.} 

The dataset was generated by scraping GrabCAD \cite{grabcad} for six different categories of 3D CAD models with related functions: \textit{Car, Engine, Robotic Arm, Airplane, Gear, and Wheel}. Functional information such as model’s name, author, description of the model, name of parts in the model, tags, likes, timestamps, and comments on the models was also extracted. The total number of models per category was 2,271 for \textit{Car}, 1,597 for \textit{Engine}, 2,013 for \textit{Robotic Arm}, 2,114 for \textit{Airplane}, 1,732 for \textit{Gear}, and 2,404 for \textit{Wheel}. These were later broken down into a testing-training split that is detailed in Section \ref{ssec:cogs}.
%\red{Could also add a table/figure/description of the actual data represented on the graphs. Can either use the same graphics as presented earlier or can use the 3D CAD models from book chapter. Can also add a textbox of an example of the metadata that was scraped from GrabCad.}
\subsection{Graph Learning Framework}
%The SGCNN architecture is depicted in Fig. \ref{fig:sgcnn}. An input graph corresponding to the dataset detailed above is constructed by [add text]. This graph is then dissected into subgraphs based on the defined query or schema. These subgraphs are then processed by a variety of operations including aggregation, graph convolutions and pooling. The objective is to correctly classify a subgraph based on its functional characteristics and those of its neighboring nodes. The main stages of the framework, aligned with our proposed cognitive digital twin framework (Graph Formation, Graph Operations, Learning Objective), are detailed in the rest of this section.
The graph learning framework proposed in \cite{wan2018future} is depicted in Fig. \ref{fig:sgcnn}. Their framework can be viewed as an application of our cognitive digital twin framework. To illustrate this, we present their proposed framework as it aligns with the different stages of our cognitive digital twin framework: Graph Formation, Graph Operations, Learning Objective. We detail these stages in their framework below.

% \begin{figure*}[!ht]
%     \centering
%     \includegraphics[width=0.9\linewidth]{GraphLearning-CDT-ManSystems/figures2/SGCNN_filler.PNG}
%     \caption{SGCNN- Architecture (filler image). \red{Request the high-res photo or powerpoint file used to generate the image. We may want to use the book version that is a bit simpler to explain.}}
%     \label{fig:sgcnn}
% \end{figure*}
\begin{figure*}[ht!]
	\centering
    %\vspace{-4em}
	\includegraphics[width=0.9\textwidth]{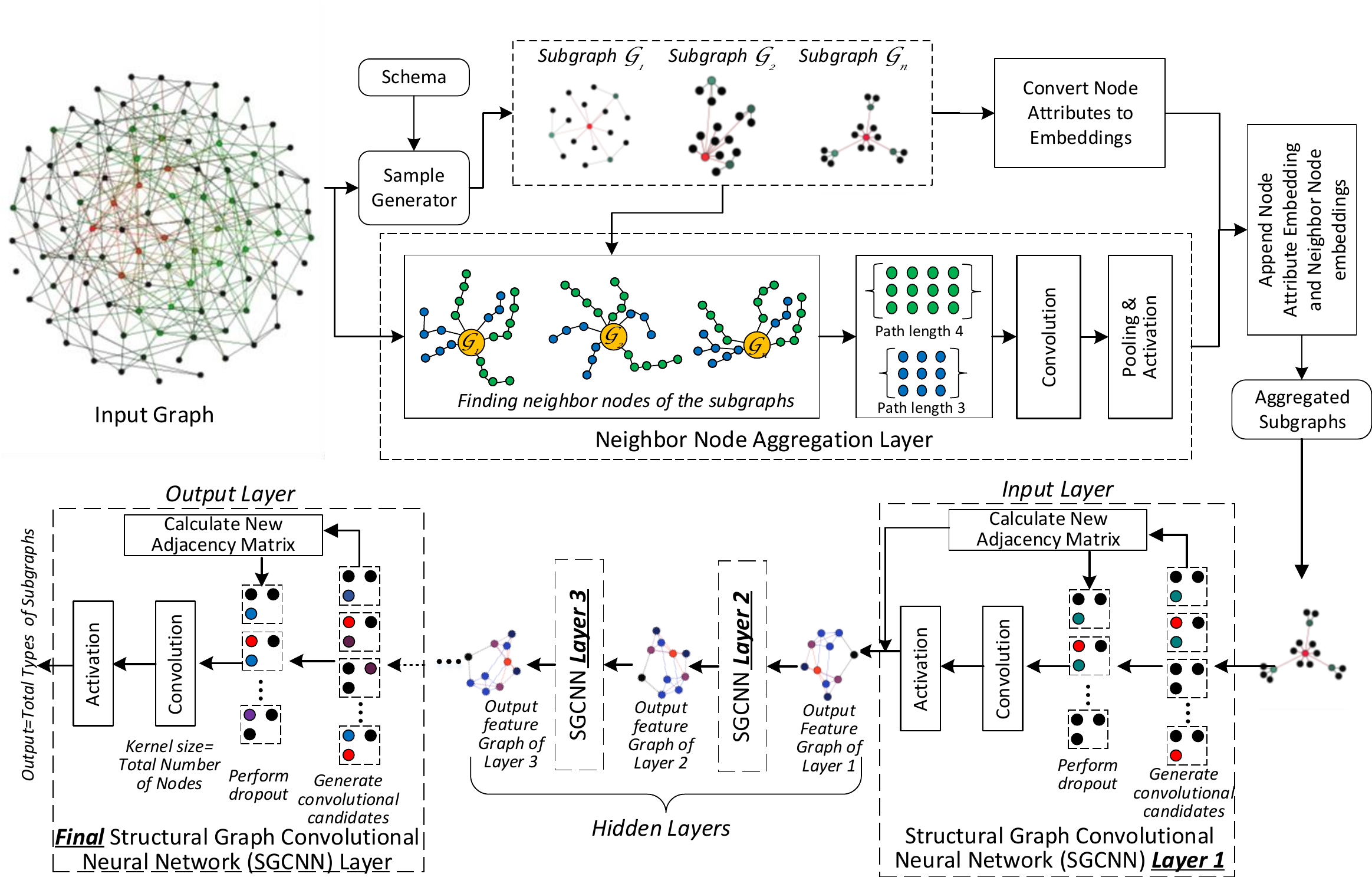}
     %\vspace{-0.5em}
	\caption{SGCNN Architecture \cite{wan2018future}.}
    %\vspace{-1em}
	\label{fig:sgcnn}
\end{figure*}

%Mathematical equations and formulae from the algorithms:
%Graph Formation: (a) knowledge graph extraction and (b) attribute embedding.
\textbf{Graph Formation}

In this stage, the input dataset is processed into graph structures based on a search query which is dependent on the target design application. Specifically, the Sample Generator generates subgraphs, $\mathcal{G}_n$, based on the queried schema of interest, which in this setting is the information from the extracted 3D CAD models. This schema is designed to extract functional information from each sample. The embedding of sample information into the feature vectors is done by the well-known \texttt{word2vec}, and the induced subgraphs are composed of the functional information for each sample. These subgraphs are aimed at capturing the meaning of the engineering design and enable functional lifting.  %\red{Details here on what the subgraphs will look like: what are the nodes and features.} The nodes of the subgraphs are .... The features of each node include .... Edges connecting subgraphs are [defined/learned]... \red{Need more time here for what is a schema. How is the graph formed - more details. Use a direct description as opposed to metadata word }

%Graph Operations: (c) Neighborhood Node Aggregation Layer and (d) SGCNN Layers
\textbf{Graph Operations}

In this stage, both graph pooling and graph convolutions are performed over the subgraphs. This involves aggregating information from neighbor nodes within a subgraph, appending this information to the original subgraphs, and then processing the subgraphs through convolutional layers. The output after these subsequent convolutions is processed through a nonlinear activation function in order to perform the classification. The final output is the probability of the various classes. The processing done during this stage enables functional lifting by abstracting the common engineering data across subgraphs in a way that they can mathematically utilized.  
The Neighborhood Node Aggregation Layer performs convolutions and pooling to aggregate information from each node's neighbors into the created subgraphs. Here, the common method \texttt{node2vec} that combines a breadth/depth first search over graphs is utilized. The feature vector for each node, $x_n$, is calculated as follows:  
\begin{equation}
    x_n = \sigma(f_{pool}(\bar{W} \circledast \bar{N}) + b) , 
\end{equation}
where $\sigma(\mathord{\cdot})$ is the activation function; $f_{pool}$ is a pooling function; $\bar{W}$ is a $1$ by $d$ trainable weight matrix with $d$ as a defined search depth; $\circledast$ represents the convolution operation; $\bar{N}$ is $n$ by $d$ matrix of the neighbor node's features (as illustrated roughly in Fig. \ref{fig:Neighbors}); and $b$ is a bias variable.

The aggregated subgraphs are then fed into the SGCNN layers which consist of subsequent convolutions, graph pooling, and non-linear activations. In order to perform the 2D convolutions over the subgraphs, an attribute matrix, $\bar{R}$, is calculated as:  
\begin{equation}
    \bar{R} = \bar{X} \circ (\bar{A}+\bar{I}),
\end{equation}
where $\bar{X}$ is the feature matrix for the subgraph (as described in Section \ref{ssec:prelims}); $\circ$ denotes the Hadamard Product of two matrices; $\bar{A}$ the adjacency matrix of the subgraph (as described in Section \ref{ssec:prelims}); and $\bar{I}$ denotes thee identity matrix. 

Before the convolutions are applied, a special graph pooling algorithm is used to down sample $\bar{R}$ to a $k$ by $k$ matrix, $\bar{R}^k$, where $k$ is the size of the convolutional kernel. The trainable $k$ by $k$ kernel matrix, $\bar{W}^k$, is passed over the subgraph as defined in the following equation:
\begin{equation}
    x^k = \phi(\bar{W}^k \circledast \bar{R}^k + b) , 
\end{equation}
where $\phi(\mathord{\cdot})$ is the non-linear activation function. This graph convolution operation is able to effectively aggregate local features for each node, in a similar manner that traditional convolutions over images are applied. The new subgraph, adjacency, and attribute matrix are then calculated using the previous subgraph and passed to the next SGCNN layer accordingly.  

%Learning Objective: (e) Subgraph Classification
\textbf{Learning Objective}
%In the setting of this example, the ultimate goal is the proper classification of subgraphs based on their functional attributes given the ground-truth labels during training

The overall graph learning objective for classification is specified by the cross-entropy loss function, $H$, over the output of the final SGCNN layer:  
\begin{equation}
    \max H, ~ H = \sum_{i \in \text{classes}} y_i log_e(\hat{y}_i) , 
\end{equation}
%\begin{equation}
%    \argmin H(Y, \hat{Y}) = \argmin \sum_{y_i \in Y, \hat{y_i} \in \hat{Y}} y_i log_e(\hat{y_i}) ,
%\end{equation}
where $y_i$ is the indicator of the ground truth validity of the label $i$ and $\hat{y}_i$ is the inferred probability of the label $i$. It is clear that when the learning objective is maximized, the predicted probabilities will closely match the ground truth labels. During training of the model, the weights of the graph pooling and convolution operations at each layer are updated. As in traditional supervised learning training approaches, the model has access to the ground truth labels to compute the loss function. During testing of the model, the performance is then evaluated against subgraphs is has not previously been trained on.  

For training and evaluation of the SGCNN architecture a train-test split of 11,304 to 2,827 samples was used. Through extensive experiments, the best SGCNN model was identified to be four layers. It was also observed that as both the number of hidden layer features and output kernel size were increased the model achieved better classification scores. It was shown that the model was able to correctly classify roughly 91\% of the subgraphs based on their functionality. We refer readers to \cite{wan2018future} for further information regarding the model structure and tuning of hyperparameters and ablation studies.

\subsection{Cognitive Capabilities}
\label{ssec:cogs}

The cognitive capabilities are utilized throughout the framework's core operations. \textit{Perception} is utilized in the representation of the data on subgraphs, and the subsequent embedding of features that are abstracted to a level for the model to perform further computations. \textit{Memory} and \textit{learning} are utilized continuously in the update and the optimization of the model parameters during the training process. In the graph operations, \textit{attention} and \textit{reasoning} are effectively employed to reuse knowledge across different samples while selectively focusing on the most relevant functional attributes. Thus, the framework implements several key cognitive functions.
%Lastly, the cognitive method of \textit{problem solving} is evident in the twin's ability to perform accurate classification

%The benefits of cognitive capabilities in this application can be observed throughout the digital twin's core operations. \textit{Perception}, \textit{memory}, and \textit{reasoning} are present during the formation of the subgraphs as the digital twin can effectively generate the appropriate relations given the input data from GrabCAD and the user defined search query/schema. \red{ Be more deliberate, be careful here. Need to flush this out better} During the operations performed over the subgraphs, \textit{attention} and \textit{learning} are effectively employed to reuse knowledge across difference samples while selective focusing on the most relevant functional attributes. Lastly, the cognitive \textit{problem solving} is evident in the twin's ability to perform accurate classification. \red{Paragraph needs refinement. Make direct connections that are straightforward - don't force fit.}

\subsection{Further Extensions}
An additional output that the graph framework could produce is the clustering of the nearest-neighbor subgraphs that serve similar design functions. This could allow the digital twin to learn from similar designs at a functional level when tasked with searching for similar products, as opposed to searching for only similar products of the same basic design. This would require a multi-task learning approach because with this additional objective the model must be trained for accurate classification and clustering. Even further, this framework can be extended for higher level cognitive functions like reasoning, decision making and problem solving by enabling a feedback loop that self-generates queries. This can potentially enable complete autonomy and fully realize the capabilities of cognitive digital twins.
%Maybe one more "head" of the model to really add some cogdt flair. Can create an image of these multiple headed pipelines along with some cognition annotations.
%Paragraph here on possible expansions such as: multi-task training and self-self learning and relations to cognition. Ties in a complete vision of future cognitive digital twins.
}%end of green.

\section{Discussion}
In this section we discuss current challenges and limitations, extensions to other areas of manufacturing, and an outlook on future research directions in the field.

\textit{Challenges:}
%\red{Update this section with the Challenges/limitation/gaps for the Application example. One major one is the rather intensive hyperaremeter tuning in turns of the model.}

There are some specific challenges with regards to the proposed framework. Formation of graphs, problem and metric specification, and hyper-parameter tuning of graph operations can require expert knowledge. \green{In the use case in Section \ref{sec:app}, extensive studies on the optimal model and hyperparameters, such as various activation functions, number of hidden features per layer, kernel size, and number of total layers were needed.} Techniques for formation of graphs are still relatively under-explored and therefore may be burdensome in some applications. \green{Additionally, there are some scenarios where the data is not easy to represent as a graph. Further abstraction of systems and/or processes may be necessary to arrive at a graph-level representation. Nonetheless, applications of graph learning approaches to these situations are still under explored.} All of these are open research challenges. 

Another notable challenge is the availability of data. Due to constraints in the sharing of propriety information, some companies may be prohibited from collaboration with others. Additionally, financial and competitive motivations may also limit progress along this direction. However, as Industry 4.0 continues to develop, the amount of available data will continue to increase as well. This will enable digital twins using data-driven approaches to become more robust and efficient over time. 
%Sentence on dynamic graphs removed
%In regards to cognition, open challenges exist in how to combine the cognitive capabilities from multiple different methods. Building a cognitive system will require high levels of coordination across multiple different complex models. The connections between the cognitive functions themselves also presents challenges in modelling.

\textit{Other Applications in Manufacturing:} 

The framework we presented can be extended to the entire life cycle of a product. During actual manufacturing, information about operations, products, logistics, and diagnostics are exchanged between various systems. This data can be modeled on graphs and graph learning can be leveraged to create new solutions to key manufacturing problems. Likewise, high-level organizational supply chains and consumer-provider supply and demands can be modelled on graphs and optimized towards specific performance objectives. Another powerful extension stems from a cognitive digital twin's ability to draw on its own past experiences (or experiences from other digital twins) to learn and adapt over time. Graph learning can incorporate temporal features, features that vary over time, which can allow a digital twin to learn from past data and observe specific time-dependent trends. This can enable iterative learning over design cycles and more intelligent design tools. Likewise, the growing areas of reconfigurable/flexible manufacturing can benefit from similar methods.

\textit{Future Research Directions:}

%We presented the search operation to concretize the abstract and generic concept of cognition in the digital twin for the design stage of the manufacturing system. However, we recognize just the search operation is far from a comprehensive approach to exploring the full benefits and potential of cognitive digital twins.
The research community has immense opportunities to contribute in the areas of mathematical representations, algorithms, tools, and methodologies for developing and using cognitive digital twins. In this context, we formulate and pose a few research questions. 

- What are the appropriate mathematical representations of digital twins that can enable the incorporation of cognitive capabilities? Examples here include differential equations, discrete-event dynamic systems, logic-based models, graph models, connectionist network models, etc. How can such models be used for simulations, state estimation, and control and decision making?

- How can high-performance computing and numerical simulation tools be leveraged to enable cognitive capabilities in digital twins? For example, can numerical simulations (along with experimental data) create large memory banks that can be used for interpreting and acting on real-time streaming data from IoT sensors? Can they be used for real-time response to changes in the manufacturing system environment?

- How can we enable public searchability of the digital twin models? More specifically, how to embed metadata in complex digital twin models (parts, processes, and systems) so that they can easily be searched over the internet during the design phase? 

- How do we make the knowledge sharing scalable in digital twin models? Scaling may fall under the scope of generalizing knowledge sharing across multiple domains. Scalability is non-trivial and is a challenge due to the complexity of cross-domain knowledge sharing. However, the digital twin models may be capable of sharing knowledge across non-overlapping domains (for example, across manufacturing systems utilizing different technologies, between the aging model and quality prediction model, etc.). 

\green{Among these future research directions, we believe that there is not one sequential path that takes precedence over others in the future development of cognitive digital twins. Different applications and studies will call for varying degrees of focus on these directions and may involve pursuing multiple directions simultaneously. Having said that, prioritizing development of strong theoretical and mathematical methods will be necessary to address many of these directions. Public searchability and scalability could be developed in tandem as they can work synergistically together to enable a proliferation of cognitive digital twin usage and prevalence.}

%\red{One short paragraph here to address the following comment: The perspective section would benefit from adding a timeline and/or a priority among the various topics.}

%\red{Focus on priority as opposed to timeline. Independent of what is written here, start with basics of cognitive digital twins, which will lead to prioritization.}

%\input{use_cases} 

\section{Conclusion}
\label{sec:conclusion}
This paper presented a comprehensive vision of future cognitive digital twins that can enable advancements in Industry 4.0. The benefits of using digital twins in manufacturing were detailed. Six core cognitive capabilities, \textit{perception, attention, memory, reasoning, problem-solving}, and \textit{learning}, were described along with their ability to influence complex manufacturing decisions and future autonomy. A novel, query-based graph learning framework for enabling cognition in digital twins was presented to fill existing research gaps in the field. We believe that to realize the full potential of cognitive digital twins, collaboration across different research communities is essential. Domain expertise in many areas will be necessary in the development of the future of manufacturing.

\section*{Acknowledgment}
This research was supported by NSF awards CMMI-1739503 and ECCS-1839429. Any opinions, findings, and conclusions or recommendations expressed in this material are those of the authors and do not necessarily reflect views of our funding agencies.

\bibliographystyle{IEEEtran}
\bibliography{IEEEabrv, main}
%\vspace{-2 cm}
\begin{IEEEbiography}[
    {\includegraphics[width=1in,height=1.25in,clip,keepaspectratio]
{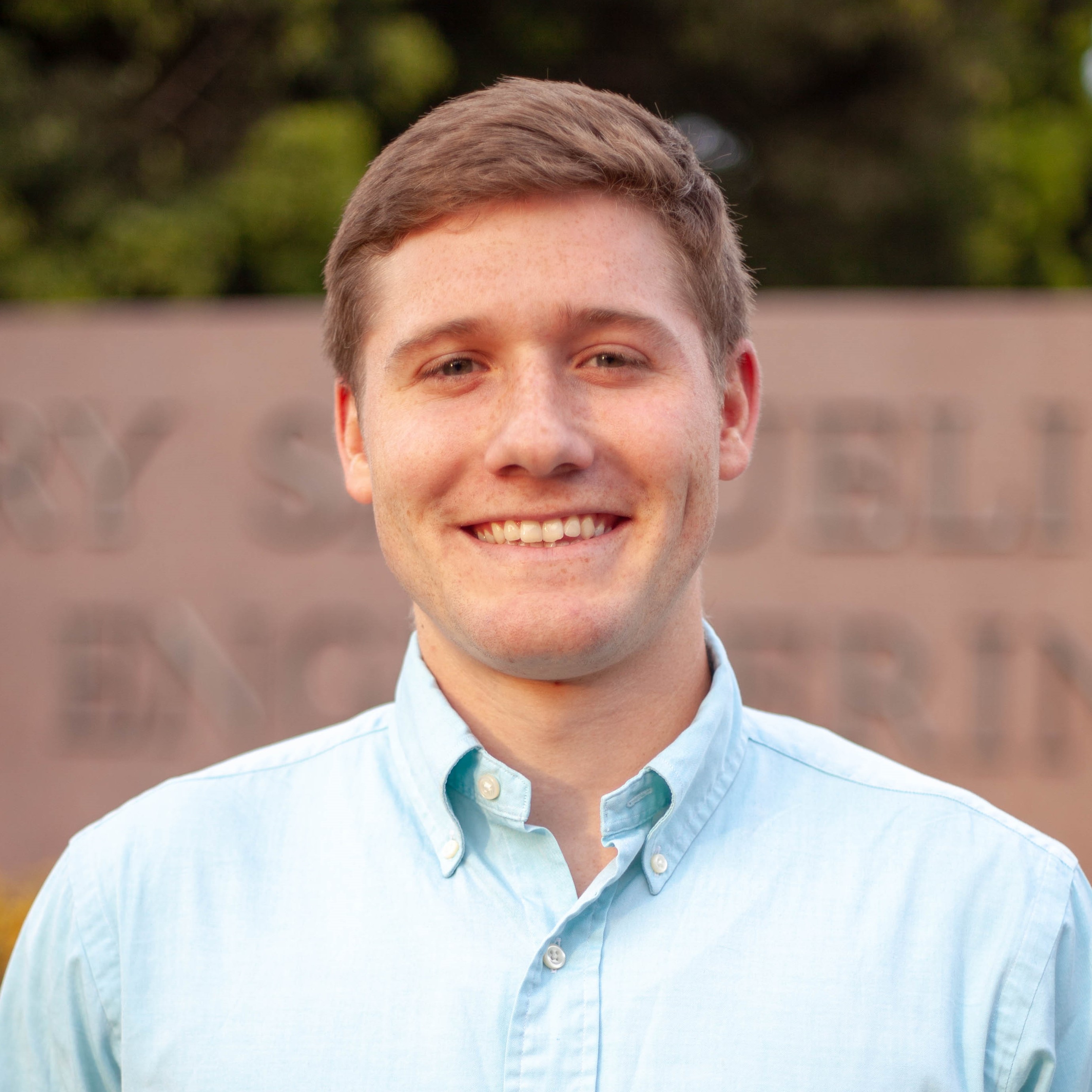}}]
{Trier Mortlock} is a Ph.D. student in the Department of Mechanical and Aerospace Engineering at the University of California, Irvine. He is a member of the Autonomous and Intelligent Cyber-Physical Systems (AICPS) Laboratory. He received his B.S. in Mechanical Engineering from the University of California, Berkeley and his M.S. in Mechanical and Aerospace Engineering from the University of California, Irvine. He has served in the United States Army Reserve since 2018, where he is presently a Cyber Operations Officer. His research interests include cyber-physical systems, autonomous vehicle security, manufacturing system security, and the intersection of machine learning and sensor fusion.
\end{IEEEbiography}
\vskip 0pt plus -1fil
\begin{IEEEbiography}[{\includegraphics[width=1in,height=1.25in,clip,keepaspectratio]{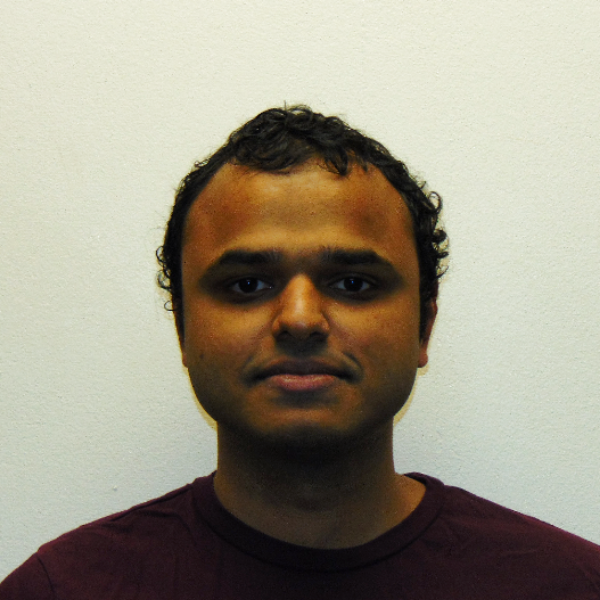}}]{Deepan Muthirayan}
is currently a Post-doctoral Researcher in the department of Electrical Engineering and Computer Science at University of California at Irvine. He obtained his Ph.D. from the University of California at Berkeley (2016) and B.Tech/M.tech degree from the Indian Institute of Technology Madras (2010). His doctoral thesis work focused on market mechanisms for integrating demand flexibility in energy systems. Before his term at UC Irvine he was a post-doctoral associate at Cornell University where his work focused on online scheduling algorithms for managing demand flexibility. His current research interests include control theory, machine learning, topics at the intersection of learning and control, online learning, online algorithms, game theory, and their application to smart systems.
\end{IEEEbiography}
\vskip 0pt plus -1fil
\begin{IEEEbiography}[
    {\includegraphics[width=1in,height=1.25in,clip,keepaspectratio]
{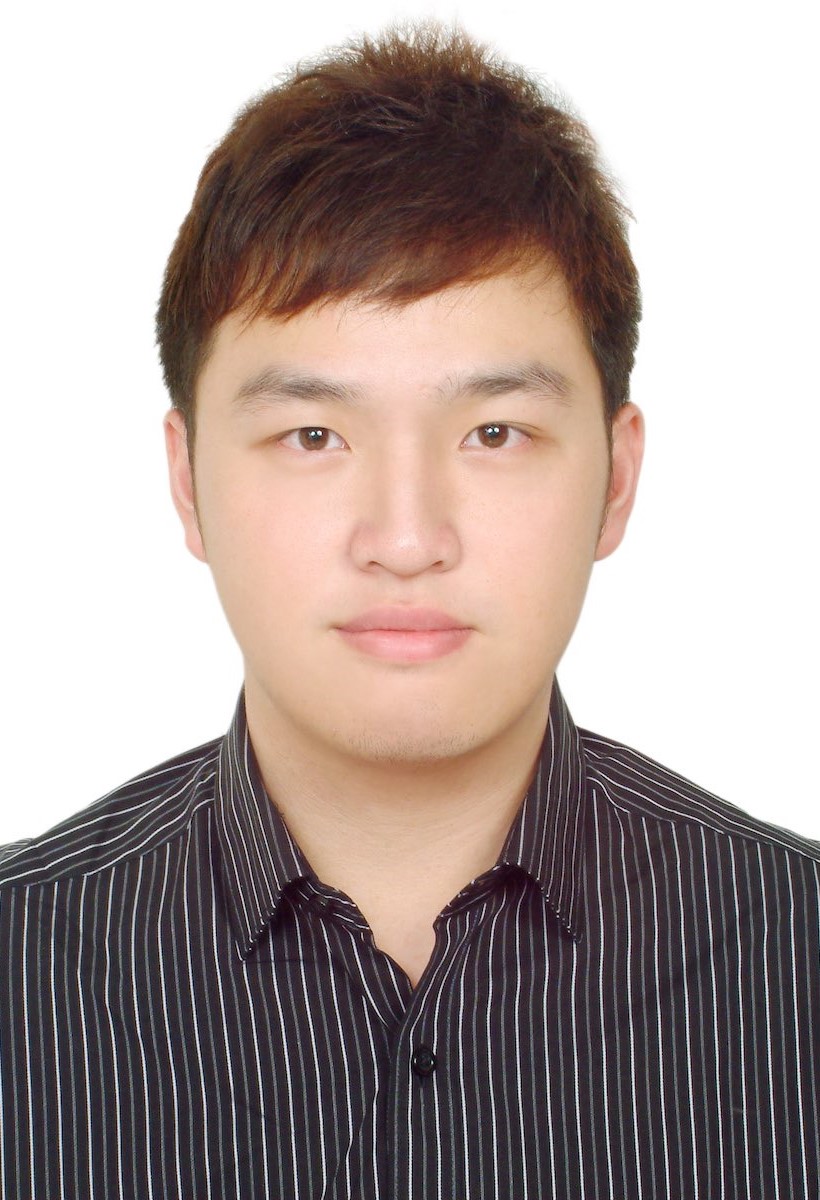}}]
{Shih-Yuan Yu} received the B.S. and M.S. degrees in Computer Science and Information Engineering from the National Taiwan University (NTU) in 2014. He worked at MediaTek for 4 years. Currently he is a Ph.D. student in the University of California, Irvine.  
Now his research interests are about design automation of embedded systems using data-driven system modeling approaches.
It covers incorporating machine learning methods to identify potential security issues in systems.
\end{IEEEbiography}
%\vspace{5 cm}
\vskip 0pt plus -1fil
\begin{IEEEbiography}
[{\includegraphics[width=1in,height=1.25in,clip,keepaspectratio]{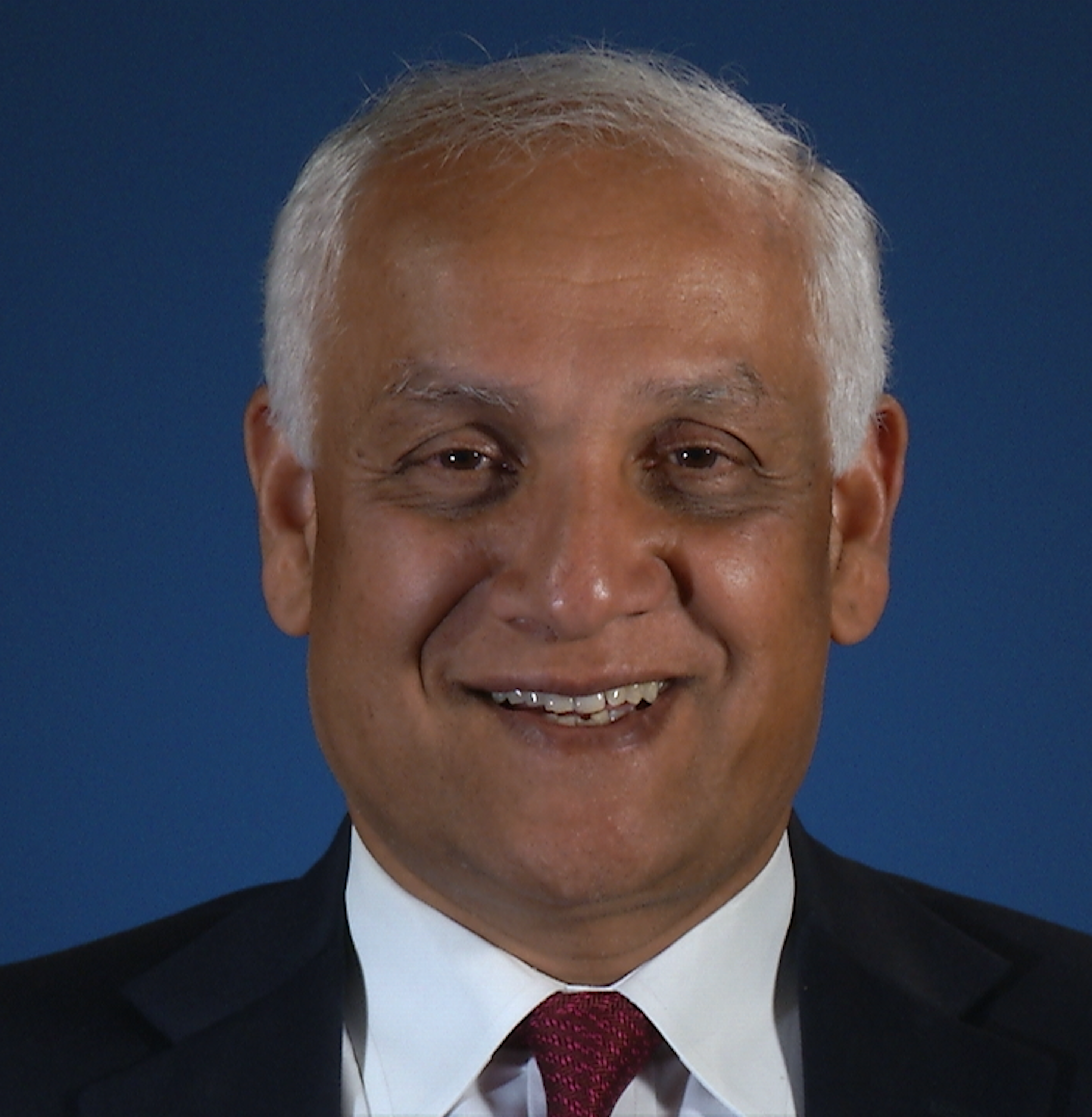}}]
{Pramod Khargonekar} received B. Tech. Degree in electrical engineering in 1977 from the Indian Institute of Technology, Bombay, India, and M.S. degree in mathematics in 1980 and Ph.D. degree in electrical engineering in 1981 from the University of Florida, respectively. He was Chairman of the Department of Electrical Engineering and Computer Science (1997-2001) and the Claude E. Shannon Professor of Engineering Science at The University of Michigan.  From 2001-2009, he was Dean of the College of Engineering and Eckis Professor of Electrical and Computer Engineering at the University of Florida (2001-2016). After serving as Deputy Director of Technology at ARPA-E (2012-13), he was appointed Assistant Director for the Directorate of Engineering (ENG) (2013-2016). Currently, he is Vice Chancellor for Research and Distinguished Professor of Electrical Engineering and Computer Science at the University of California, Irvine. His research and teaching interests are centered on theory and applications of systems and control. 
% {Pramod Khargonekar} received B. Tech. Degree in electrical engineering in 1977 from the Indian Institute of Technology, Bombay, India, and M.S. degree in mathematics in 1980 and Ph.D. degree in electrical engineering in 1981 from the University of Florida, respectively. He was Chairman of the Department of Electrical Engineering and Computer Science from 1997 to 2001 and also held the position of Claude E. Shannon Professor of Engineering Science at The University of Michigan.  From 2001 to 2009, he was Dean of the College of Engineering and Eckis Professor of Electrical and Computer Engineering at the University of Florida till 2016. After serving briefly as Deputy Director of Technology at ARPA-E in 2012-13, he was appointed by the National Science Foundation (NSF) to serve as Assistant Director for the Directorate of Engineering (ENG) in March 2013, a position he held till June 2016. Currently, he is Vice Chancellor for Research and Distinguished Professor of Electrical Engineering and Computer Science at the University of California, Irvine. His research and teaching interests are centered on theory and applications of systems and control. He has received numerous honors and awards including IEEE Control Systems Award, IEEE Baker Prize, IEEE CSS Axelby Award, NSF Presidential Young Investigator Award, AACC Eckman Award, and is a Fellow of IEEE, IFAC, and AAAS.
\end{IEEEbiography}
\vskip 0pt plus -1fil
\begin{IEEEbiography}
[{\includegraphics[width=1in,height=1.25in,clip,keepaspectratio]
{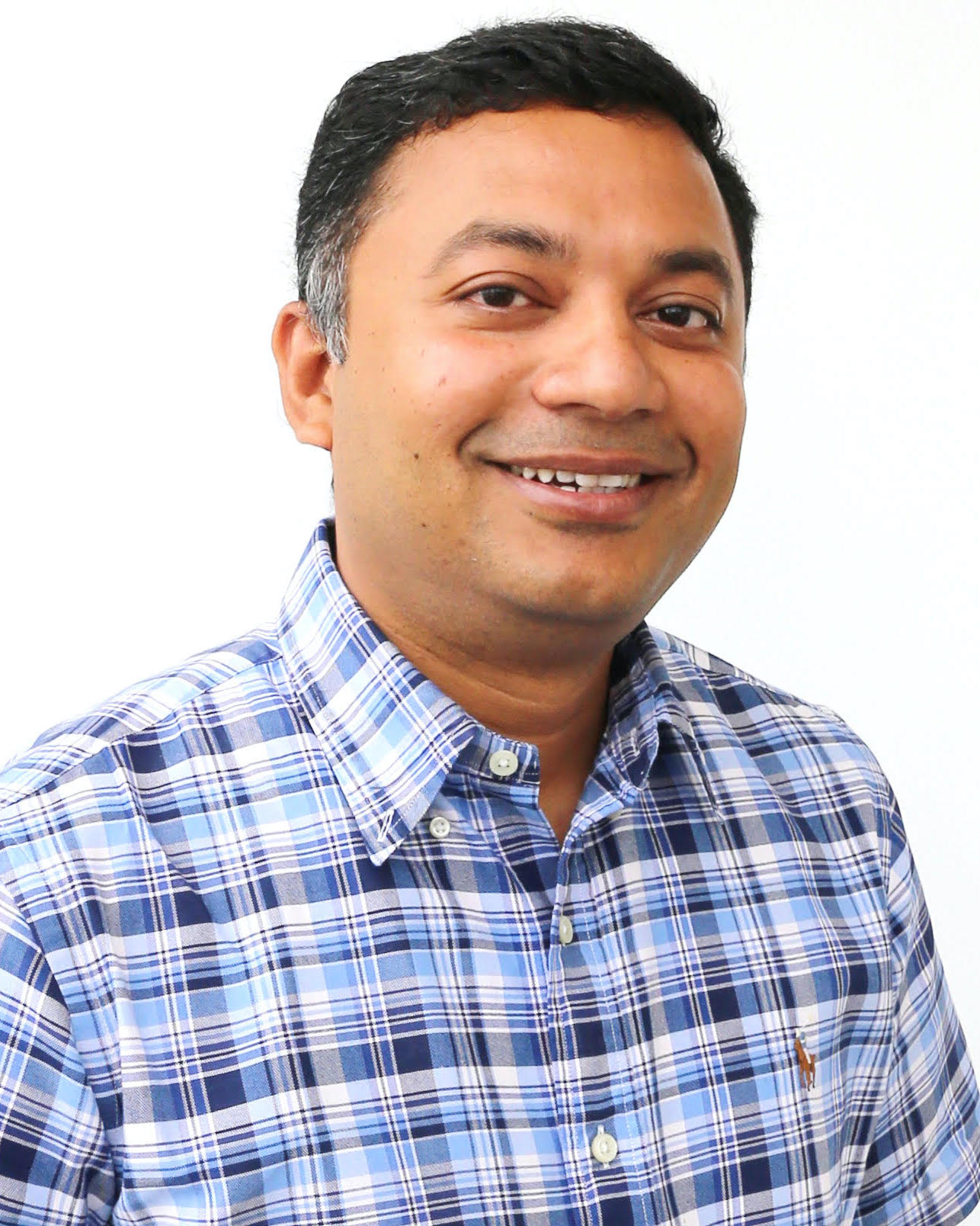}}]
{Mohammad Abdullah Al Faruque} (M’06, SM’15) received his B.Sc. degree in Computer Science and Engineering (CSE) from Bangladesh University of Engineering and Technology (BUET) in 2002, and M.Sc. and Ph.D. degrees in Computer Science from Aachen Technical University and Karlsruhe Institute of Technology, Germany in 2004 and 2009, respectively.
He is an Associate Professor at University of California, Irvine (UCI) and Director of the Embedded and Cyber-Physical Systems Lab. He was with Siemens Corporate Research and Technology in Princeton, NJ as a Research Scientist. His current research is focused on the system-level design of embedded and Cyber-Physical-Systems (CPS) with special interest in low-power design, CPS security, data-driven CPS design, etc.
Besides many other awards, he is the recipient of the School of Engineering Mid-Career Faculty Award for Research 2019, the IEEE Technical Committee on Cyber-Physical Systems Early-Career Award 2018, and the IEEE CEDA Ernest S. Kuh Early Career Award 2016. 
% {Mohammad Abdullah Al Faruque} (M’06, SM’15) received his B.Sc. degree in Computer Science and Engineering (CSE) from Bangladesh University of Engineering and Technology (BUET) in 2002, and M.Sc. and Ph.D. degrees in Computer Science from Aachen Technical University and Karlsruhe Institute of Technology, Germany in 2004 and 2009, respectively.
% He is currently with the University of California Irvine (UCI) as an Associate Professor and Directing the Embedded and Cyber-Physical Systems Lab. He served as an Emulex Career Development Chair from October 2012 till July 2015. Before, he was with Siemens Corporate Research and Technology in Princeton, NJ as a Research Scientist. His current research is focused on the system-level design of embedded and Cyber-Physical-Systems (CPS) with special interest in low-power design, CPS security, data-driven CPS design, etc.
% He is an ACM senior member. He is the author of 2 published books. Besides many other awards, he is the recipient of the School of Engineering Mid-Career Faculty Award for Research 2019, the IEEE Technical Committee on Cyber-Physical Systems Early-Career Award 2018, and the IEEE CEDA Ernest S. Kuh Early Career Award 2016. He is also the recipient of the UCI Academic Senate Distinguished Early-Career Faculty Award for Research 2017 and the School of Engineering Early-Career Faculty Award for Research 2017. Besides 120+ IEEE/ACM publications in the premier journals and conferences, he holds 9 US patents.
\end{IEEEbiography}
\end{document}